\documentclass[journal, onecolumn]{IEEEtran}
\IEEEoverridecommandlockouts
\pdfoutput=1

\usepackage[utf8]{inputenc}
\usepackage[T1]{fontenc}
\usepackage{graphicx}
\graphicspath{{./figs/}}
\DeclareGraphicsExtensions{.pdf}
\usepackage{xcolor}

\usepackage[cmex10]{amsmath}
\usepackage{amsthm,amssymb,amsfonts,mathrsfs,bm,mathtools}
\mathtoolsset{centercolon=true}

\DeclareMathAlphabet{\mathpzc}{OT1}{pzc}{m}{it}

\usepackage{newfloat}
\usepackage[caption=false,labelsep=quad,indention=1pt]{subfig}
\usepackage{setspace}
\usepackage[inline]{enumitem}
\usepackage{algorithm}
\usepackage{algpseudocode}
\usepackage{soul}
\usepackage{xcolor}

\usepackage{etoolbox}
\makeatletter
\expandafter\patchcmd\csname\string\algorithmic\endcsname{\itemsep\z@}{\itemsep=.6ex plus1pt}{}{}
\makeatother
\AtBeginEnvironment{algorithmic}{\setstretch{1}}
\usepackage{float}
\usepackage[hyphens]{url}
\usepackage[noadjust,sort,compress]{cite}
\usepackage{xspace}

\usepackage{pgfplots}
\pgfplotsset{compat=newest}
\usepackage[bwr]{callouts}
\usepackage{tikz}
\pgfplotsset{select coords between index/.style 2 args={
    x filter/.code={
        \ifnum\coordindex<#1\def\pgfmathresult{}\fi
        \ifnum\coordindex>#2\def\pgfmathresult{}\fi
    }
}}

\newcommand{\Complex}{\mathbb{C}}

\newcommand\vect[1]{\mathbf{#1}}

\DeclarePairedDelimiterX\Set[1]{\lbrace}{\rbrace}%
{  #1 }
\DeclarePairedDelimiterX\innerp[2]{\langle}{\rangle}{#1
  \mathop{}\delimsize\vert\mathop{} #2}
\DeclarePairedDelimiterX\norm[1]\lVert\rVert{\ifblank{#1}{\:\cdot\:}{#1}}

\DeclareMathOperator{\diag}{diag}

\DeclareMathOperator*{\Argmin}{arg\,min}

\DeclareMathOperator{\Vect}{vec}

\newcommand*{\hermconj}{\mathsf{H}}

\let\oldsqrt\sqrt
\def\sqrt{\mathpalette\DHLhksqrt}
\def\DHLhksqrt#1#2{%
\setbox0=\hbox{$#1\oldsqrt{#2\,}$}\dimen0=\ht0
\advance\dimen0-0.2\ht0
\setbox2=\hbox{\vrule height\ht0 depth -\dimen0}%
{\box0\lower0.4pt\box2}}

\newtheoremstyle{kostasstyle}
{2pt}
{2pt}
{\normalfont}
{0pt}
{\bfseries}
{.}
{1ex}
{}

\theoremstyle{kostasstyle}

\newtheorem{assumption}{Assumption}

\makeatletter

\newcommand*{\ie}{%
  \@ifnextchar{,}%
  {\textit{i.e.}}%
  {\textit{i.e.,}\@\xspace}%
}
\newcommand*{\eg}{%
  \@ifnextchar{,}%
  {\textit{e.g.}}%
  {\textit{e.g.,}\@\xspace}%
}
\newcommand*{\etc}{%
  \@ifnextchar{.}%
  {\textit{etc}}%
  {\textit{etc.}\@\xspace}%
}
\newcommand*{\etal}{%
  \@ifnextchar{.}%
  {\textit{et al}}%
  {\textit{et al.}\@\xspace}%
}
\newcommand*{\cf}{%
  \@ifnextchar{.}%
  {\textit{cf}}%
  {\textit{cf.}\@\xspace}%
}
\newcommand*{\aka}{%
  \@ifnextchar{,}%
  {\textit{a.k.a.}}%
  {\textit{a.k.a.}\@\xspace}%
}
\makeatother

\makeatletter
\def\endthebibliography{%
  \def\@noitemerr{
  \@latex@warning{Empty `thebibliography' environment}}%
  \endlist
}
\newcommand*{\balancecolsandclearpage}{%
  \close@column@grid
  \clearpage
  \twocolumngrid
}
\makeatother

\renewcommand\footnotemark{}
\MakeRobust{\eqref}
\MakeRobust{\cf}
\usepackage{textcomp}
\def\BibTeX{{\rm B\kern-.05em{\sc i\kern-.025em b}\kern-.08em
    T\kern-.1667em\lower.7ex\hbox{E}\kern-.125emX}}
    
\begin{document}

\title{Kernel Bi-Linear Modeling for Reconstructing\\ Data on Manifolds: The Dynamic-MRI Case
\thanks{This work was supported in part by the NSF grant 1718796.}
}

\author{
    \IEEEauthorblockN{
        Gaurav~N.~Shetty\IEEEauthorrefmark{1},
        Konstantinos~Slavakis\IEEEauthorrefmark{2}, 
        Ukash~Nakarmi\IEEEauthorrefmark{3},
        Gesualdo~Scutari\IEEEauthorrefmark{4}, and
        Leslie~Ying\IEEEauthorrefmark{5}}
    \IEEEauthorblockA{
        Depts. of \IEEEauthorrefmark{1}\IEEEauthorrefmark{2}\IEEEauthorrefmark{5}
        Electrical Engineering and 
        \IEEEauthorrefmark{5}Biomedical Engineering, 
        University at Buffalo, Buffalo, NY 14260, USA\\
        \IEEEauthorrefmark{3}Depts.\ of Electrical Engineering and Radiology, 
        Stanford University, Stanford, CA 94305, USA\\
        \IEEEauthorrefmark{4}Dept. of Industrial Engineering, 
        Purdue University, West Lafayette, IN 47907-2023, USA\\
    Emails:  \{\IEEEauthorrefmark{1}gauravna,
            \IEEEauthorrefmark{2}kslavaki,
            \IEEEauthorrefmark{5}leiying
            \}@buffalo.edu,
            \IEEEauthorrefmark{3}nakarmi@stanford.edu,
            \IEEEauthorrefmark{4}gscutari@purdue.edu}
}

\maketitle

\begin{abstract}
This paper establishes a kernel-based framework for
reconstructing data on manifolds, tailored to fit the dynamic-(d)MRI-data
recovery problem. The proposed methodology exploits simple
tangent-space geometries of manifolds in reproducing kernel
Hilbert spaces, and follows classical kernel-approximation
arguments to form the data-recovery task as a bi-linear inverse
problem. Departing from mainstream approaches, the proposed
methodology uses no training data, employs no graph Laplacian
matrix to penalize the optimization task, uses no costly (kernel)
preimaging step to map feature points back to the input space, and utilizes
complex-valued kernel functions to account for k-space data. The
framework is validated on synthetically generated dMRI data,
where comparisons against state-of-the-art schemes highlight the
rich potential of the proposed approach in data-recovery
problems.
\end{abstract}


\section{Introduction}\label{sec:intro}


High-quality and artifact-free image reconstruction is a
perennial task in a plethora of imaging technologies, such as
dMRI; a high-fidelity visualization technology with
widespread applications in cardiac-cine, dynamic-contrast
enhanced and neuro-imaging~\cite{Liang.Lauterbur.book, Liang.Lauterbur.dMRI.94}. Due to various 
physiological constraints, MR scanners are usually slow in data acquisition  
and unable to keep up with organ motions or
fluid flow~\cite{Liang.Lauterbur.book}. Moreover, there is a
constant need to speed up the data collection process to make it inexpensive 
and to cause less patient discomfort. The unpleasant end result is the availability 
of a limited number of measurements/data, which leads in turn to distorted and aliasing
image reconstructions~\cite{Liang.Lauterbur.book, Liang.Lauterbur.dMRI.94}.

Naturally, a lot of research focuses on reconstruction 
algorithms that yield high-fidelity image series given the usually 
inadequate number of measurements from MR scanners. 
Recent years have seen a paradigm shift to artificial-intelligence (AI) approaches in 
image reconstruction, such as convolutional neural networks 
(CNNs)~\cite{liang2020deep, schlemper2018deep} and generative adversarial networks
(GANs)~\cite{mardani2018deep, hammernik2018learning}. However, these techniques
rely on the availability of large number of training data and a time consuming 
training phase preceding the reconstruction task.


In contrast to AI approaches, this paper focuses on methods that
do not use any training data, but explore and exploit instead latent
geometries of the observed data. Among such methods, a mainstream
approach is compressed sensing~\cite{lustig2006kt,otazo2010combination, Jung.07, 
zhao2012pssparse, wen2017frist}. 
For instance,~\cite{zhao2012pssparse} estimates first 
a temporal basis from image series via singular-value decomposition, followed
by the estimation of a spatial basis through a sparsity inducing convex optimization
task.


Manifold-learning techniques, identifying intrinsic manifold
geometries beneath raw data for data recovery, have found
applications in face recognition~\cite{jiang2013locality,
zhang2016sparse}, as well as dMRI~\cite{poddar2016dynamic}. 
{Use of local SVD/PCA in data modeling has gathered also interest in applications not necessarily related to MRI~\cite{little2017multiscale}.}
A graph Laplacian matrix, capturing relations within data clouds,
is used as a regularizer in a convex recovery problem
in~\cite{poddar2016dynamic}. Motivated by the success of kernel
methods~\cite{scholkopf2001learning} in extending linear
models to non-linear counterparts, the MRI community has also
seen the introduction of kernel methods in MRI data recovery
tasks~\cite{poddar2018recovery, nakarmi2017kernel,
arif2019accelerated}. Notwithstanding, such kernel-based
approaches require explicit and often costly
(kernel) preimaging steps, and leave the complex-valued nature of the
k-space data underused.

The present study builds on~\cite{shetty2019bi} to establish its
generalization towards manifold geometries in reproducing kernel
Hilbert spaces (RKHSs). To promote a computationally efficient
scheme, a small set of landmark points is chosen to describe
concicely the data cloud in the input space, and then mapped
non-linearly to the RKHS feature space. Driven by the continuity
of the mapping (kernel) function, it is hypothesized that the
mapped feature points lie on or close to a smooth unknown and
low-dimensional manifold. Departing from mainstream
manifold-learning approaches, where the graph Laplacian operator
is used as a regularizer in an inverse problem, e.g.,
\cite{poddar2016dynamic}, this study relies heavily on the
concept of tangent spaces of smooth manifolds to establish a
local and data-adaptive method of revealing affine relations
among data points. These affine relations, which take place in
the RKHS feature space, amount to non-linear operations in the
original input space, yielding thus better data approximations
according to kernel-method arguments~\cite{scholkopf2001learning,
wand1994kernel}. Costly preimaging operations to solve for
vectors in the original input space are avoided via a bi-linear
modeling approach. The framework employs also a dimensionality
reduction module to impose a low-rank structure to the
data. Furthermore, terms capitalizing on the periodicity that
exists often in dMRI are also introduced to achieve high-fidelity
image reconstructions. Finally, recently developed convex and
non-convex minimization techniques are employed to solve the
resultant recovery tasks. It is worth stressing here the points
that distinguish the proposed framework from prior art:
\begin{enumerate*}[label=\textbf{\roman*})]
\item No training data are used;
\item in contrast to mainstream manifold-learning techniques,
no graph Laplacian matrix is used to penalize the optimization task;
\item bi-linear terms are used to approximate data;
\item no costly (kernel) preimaging step is necessary to map feature points back to the input space; and
\item complex-valued kernel functions are defined to account for the k-space data.
\end{enumerate*}
Preliminary numerical tests on synthetically generated
high-dimensional dMRI phantom data and comparisons against
state-of-the-art methodologies underline the rich potential of
the method for high-quality data recovery.


\begin{figure}[t]
  \centering
  \subfloat[\label{fig:DataCube}]{\includegraphics[width = .25\linewidth]{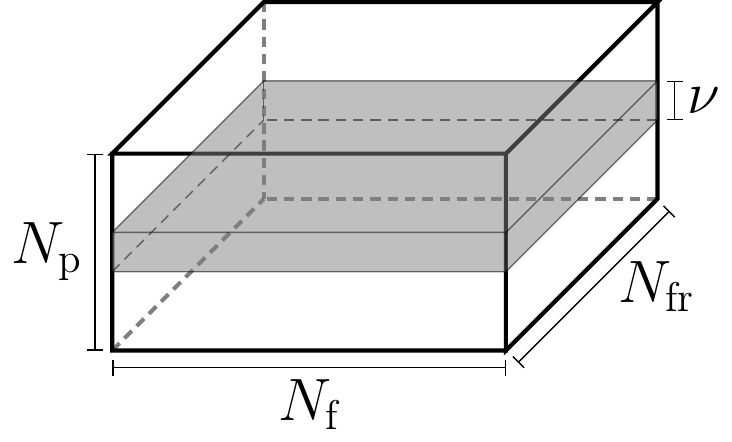}} 
  \quad
  \subfloat[\label{fig:sampling.strategy.cartesian}] {\includegraphics[width =
    .25\linewidth]{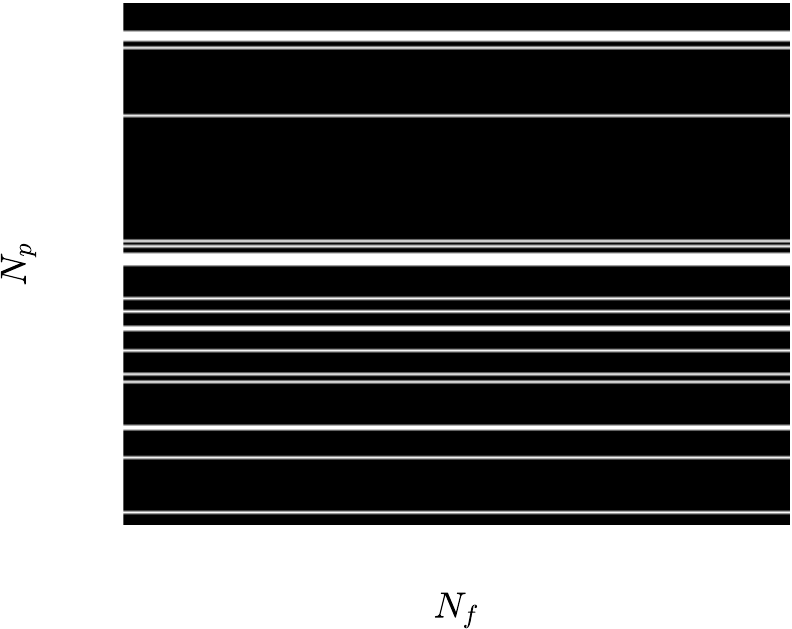}}
  \quad
  \subfloat[\label{fig:data.manifold}] {\includegraphics[width =
    0.42\linewidth]{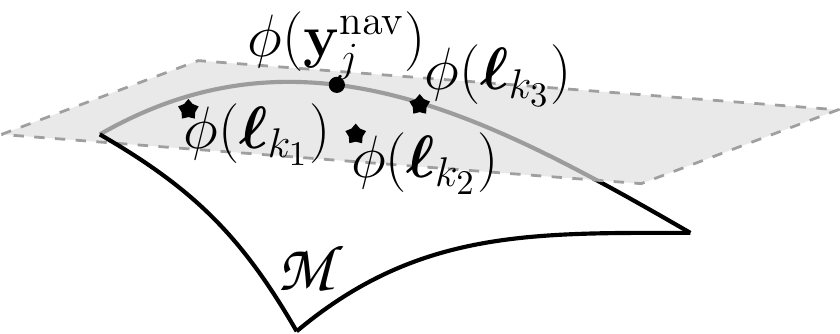}}
  \caption{
    \protect\subref{fig:DataCube} The $N_{\text{p}} \times N_{\text{f}} 
    \times N_{\text{fr}}$ (k,t)-space. ``Navigator (pilot) data'' comprise 
    the gray-colored $\nu \times N_{\text{f}} \times N_{\text{fr}}$ 
    area of the (k,t)-space ($\nu\ll N_{\text{p}}$).
    \protect\subref{fig:sampling.strategy.cartesian} 
    k-space with 1-D Cartesian sampling pattern.
    \protect\subref{fig:sampling.strategy.cartesian}
    In the feature space $\mathcal{H}_\kappa$, 
    mapped landmark points $\{\bm{\phi}(\bm{\ell}_{k_i})\}_{i=1}^3$
    are affinely combined to describe $\bm{\phi}(\vect{y}_j^\text{nav})$. 
    All possible affine combinations of $\{\bm{\phi}(\bm{\ell}_{k_i})\}_{i=1}^3$ 
    are represented by the gray-coloured area.} 
    \label{fig:ktspace}
\end{figure}

\section{DMRI-Data Description \& Representation}\label{sec:data.rep}

An  MR image $\bm{\mathcal{X}} \in \Complex^{N_\text{p} \times N_\text{f}}$ 
($\Complex$ is a set  of all complex numbers) is observed via 
$\bm{\mathcal{Y}} \in \Complex^{N_\text{p} \times N_\text{f}}$ at discrete k-space 
(or frequency domain) locations, spanning an area of $N_\text{p} \times N_\text{f}$ 
where $N_\text{p}$ and $N_\text{f}$ stand for number of phase and frequency encoding lines,
respectively. Measurements $\bm{\mathcal{Y}}$ can be viewed as the two-dimensional (discrete) 
Fourier transform of $\bm{\mathcal{X}}$: $\bm{\mathcal{Y}} = \mathcal{F}(\bm{\mathcal{X}})$~\cite{Liang.Lauterbur.book}. 
Without any loss of generality, it is assumed that the ``low-frequency'' 
part of $\bm{\mathcal{Y}}$ is located around the center of the $N_\text{p} \times N_\text{f}$ 
area. In dMRI, a temporal dimension is added to yield
$\bm{\mathcal{Y}} \in \Complex^{N_\text{p} \times N_\text{f} \times N_\text{fr}}$
where $N_\text{fr}$ represents the number of frames collected over time  
(Fig.~\ref{fig:DataCube}), resulting in the augmented 
(k,t)-space. The (k,t)-space can be seen as the collection of k-space measurements 
$\bm{\mathcal{Y}}_i = \mathcal{F}(\bm{\mathcal{X}}_j)$,
$\forall j \in \{1, \dots, N_\text{fr}\}$.

In practice, due to physical and physiological 
phenomena, k-space measurements are missing, causing severe 
\textit{undersampling} of the k-space, which in turn results in distorted, artifact-induced 
reconstructions of $\bm{\mathcal{X}}_j = \mathcal{F}^{-1}(\bm{\mathcal{Y}}_j)$~\cite{Liang.Lauterbur.book}. 
The present framework's objective is to 
reconstruct images from 
limited k-space measurements but free of aliasing effects and distortions. 
These limited k-space measurements are usually made 
along predefined trajectories for efficient acquisition of MR data~\cite{Liang.Lauterbur.dMRI.94}. 
One such trajectory is the one-dimesnional (1D) Cartesian sampling, exhibited in 
Fig.~\ref{fig:sampling.strategy.cartesian}, where measurements are recorded 
only along the ``white'' lines and the ``black'' space is assumed to be 
zero-filled. To this end, the proposed framework considers the availability of 
these heavily sampled central k-space data, coined \textit{navigator} data which are 
usually a small number $\nu (\ll N_\text{p})$ of phase encoded  lines 
(``gray-coloured region'' in Fig.~\ref{fig:DataCube}), and 
the highly (pseudo-randomly) undersampled high frequency region as in 
\cite{nakarmi2017kernel, nakarmi2018mls, poddar2018recovery, shetty2019bi}.

To facilitate processing, (k,t)-space data are 
vectorised. More specifically, $\Vect(\bm{\mathcal{Y}}_j)$ operation stacks 
one column of $\bm{\mathcal{Y}}_j$ below the other to form a $N_\text{k} \times 1$ 
vector $\vect{y}_j = \Vect(\bm{\mathcal{Y}}_j)$, 
where $N_{\text{k}} := N_{\text{p}}N_{\text{f}}$. 
To avoid notation clutter, $\mathcal{F}$ still denotes the two-dimensional 
(discrete) Fourier transform even when applied to vectorized versions of image frames:
$\mathcal{F}[\Vect(\bm{\mathcal{X}}_j)] := \Vect[\mathcal{F}(\bm{\mathcal{X}}_j)]=
\Vect(\bm{\mathcal{Y}}_j)$. All vectorized k-space frames are gathered in the
$N_{\text{k}}\times N_{\text{fr}}$ matrix $\vect{Y} := [\vect{y}_1, \vect{y}_2, \ldots, \vect{y}_{N_{\text{fr}}}]$ so that the vectorized image-domain data are
$\vect{X} := \mathcal{F}^{-1}(\vect{Y}) := [\mathcal{F}^{-1}(\vect{y}_1),
\mathcal{F}^{-1}(\vect{y}_2), \ldots, \mathcal{F}^{-1}(\vect{y}_{N_{\text{fr}}})]$.
The navigator data of the $j$th k-space frame 
(\cf~Fig.~\ref{fig:DataCube}), are gathered into a $\nu N_{\text{f}}\times 1$ vector $\vect{y}_j^{\text{nav}}$. All navigator data comprise the $\nu N_{\text{f}}\times N_{\text{fr}}$ matrix $\vect{Y}_{\text{nav}} := 
[\vect{y}_1^{\text{nav}}, \vect{y}_2^{\text{nav}}, \ldots, 
\vect{y}_{N_{\text{fr}}}^{\text{nav}}]$.

\section{Data Modeling}\label{sec:background}

The navigator data $\vect{Y}_\text{nav}$ carries useful spatio-temporal 
information in the (k,t) space. However, $N_\text{fr}$ tends to be usually 
large and to promote parsimonious data representations, it is desirable 
to extract a subset $\{\bm{\ell}_k\}_{k=1}^{N_{\ell}} \subset 
\{\vect{y}^\text{nav}_j\}_{j=1}^{N_\text{fr}}$ (coined \textit{landmark points};
$N_\ell \leq N_\text{fr}$) which concisely describes the data cloud
$\{\vect{y}^\text{nav}_j\}_{j=1}^{N_\text{fr}}$. Stacking 
$\{\bm{\ell}_k\}_{k=1}^{N_{\ell}}$ as columns in a $\nu N_\text{f} \times 
N_\text{fr}$ matrix, $\bm{\Lambda} := [\bm{\ell}_1, \bm{\ell}_2, \ldots, 
\bm{\ell}_{N_{\ell}}]$ is defined. To reveal potential non-linear spatio-temporal dependencies in $\vect{Y}_\text{nav}$,  
embedding into a high dimensional (potentially infinite) complex RKHS
$\mathcal{H}_\kappa$ \textit{(feature space)} is pursued. The feature map 
$\bm{\phi}: \Complex^{\nu N_\text{f}} \to 
\mathcal{H}_\kappa: \vect{y}_j^\text{nav} \mapsto \bm{\phi}(\vect{y}_j^\text{nav}):= \kappa(\cdot, \vect{y}_j^\text{nav})$ 
is defined via the reproducing 
kernel $\kappa(\cdot, \cdot)$~\cite{scholkopf2001learning}. The celebrated 
\textit{kernel trick}/\ $\kappa(\vect{y}_i^\text{nav}, \vect{y}_j^\text{nav}) = 
\innerp{\bm{\phi}(\vect{y}_i^\text{nav})}{\bm{\phi}(\vect{y}_j^\text{nav})}
\eqqcolon \bm{\phi}(\vect{y}_i^\text{nav})^\hermconj \bm{\phi}(\vect{y}_j^\text{nav})$, facilitates
processing~\cite{scholkopf2001learning}, where $\innerp{\cdot}{\cdot}$ is the inner product associated with $\mathcal{H}_\kappa$, and $\hermconj$, which denotes the Hermitian transposition of 
vectors/matrices, is used here to simplify notations.

All landmark points, which are mapped to $\mathcal{H}_\kappa$, can be stacked into matrix
$\bm{\Phi}(\vect{\Lambda}) := [\bm{\phi}(\bm{\ell}_1), \bm{\phi}(\bm{\ell}_2), \ldots, 
\bm{\phi}(\bm{\ell}_{N_{\ell}})]$. The kernel matrix $\vect{K} = [K_{ij} := \kappa(\bm{\ell}_i, \bm{\ell}_j)]$ can be written concisely as 
$\vect{K} \coloneqq \bm{\Phi}(\vect{\Lambda})^\hermconj \bm{\Phi}(\vect{\Lambda})$. A range of kernel functions is available for complex RKHS, such as the Gaussian 
$\kappa(\bm{\ell}_i, \bm{\ell}_j) = \text{exp}(-\gamma \| \bm{\ell}_i - \bm{\ell}_j^*\|^2)$
($*$ represents vector/matrix conjugation) and 
the polynomial kernel 
$\kappa(\bm{\ell}_i, \bm{\ell}_j) = ({\bm{\ell}_i}^\hermconj\bm{\ell}_j + c)^r$~\cite{boloix2015complex}. The following assumption 
imposes a structure on $\bm{\phi}(\vect{y}^\text{nav}_j)$; a condition often met 
in manifold-learning approaches~\cite{Saul.Roweis.03}.
\begin{assumption}\label{as:smooth.manifold}
    Data $\{\bm{\phi}(\vect{y}^\text{nav}_j)\}_{j=1}^{N_\text{fr}}$ lie on 
    a smooth low-dimensional manifold $\mathpzc{M}$~\cite{Tu.book.08} in 
    the high dimensional RKHS $\mathcal{H}_\kappa$ (\cf Fig.~\ref{fig:data.manifold}).
\end{assumption}

Following As.~\ref{as:smooth.manifold}, the concept of tangent spaces of smooth 
manifolds can be employed to approximate each $\bm{\phi}(\vect{y}_j^\text{nav})$ as an affine
combination of its neighbouring points in the feature space. 
Motivated by Fig.~\ref{fig:data.manifold}, where $\vect{y}_j^\text{nav}$ 
is described by the ``gray coloured'' area depicting
all possible combinations of $\{\bm{\phi}(\bm{\ell}_{k_1}), \bm{\phi}(\bm{\ell}_{k_2}),
\bm{\phi}(\bm{\ell}_{k_3})\}$, the existence of an $N_\ell \times 1$ vector 
$\bm{\beta}_j \in \Complex^{N_\ell}$ that approximates 
$\bm{\phi}(\vect{y}_j^\text{nav}) \approx \bm{\Phi}(\vect{\Lambda})\bm{\beta}_j$ is assumed. 
Since affine combinations are desired, constraint 
$\vect{1}_{N_{\ell}}^{\intercal} \bm{\beta}_j = 1$ is imposed where $\vect{1}_{N_{\ell}}$ 
is an all-one $N_\ell \times 1$ vector and $\intercal$ denotes vector/matrix 
transposition. Since $\bm{\phi}(\vect{y}_j^\text{nav})$ is approximated 
by points in its close vicinity, $\bm{\beta}_j$ can be considered sparse. 
Motivated also by standard kernel-approximation arguments~\cite{wand1994kernel}, 
a measurement $y$ of the k-space data vector $\vect{y}_j$ is modeled here via
$y \approx \innerp{\sum_{i=1}^{N_\ell} \alpha_i \bm{\phi}(\bm{\ell}_i)}{\bm{\phi}(\vect{y}_j^\text{nav})} 
= \innerp{\bm{\Phi}(\vect{\Lambda})\bm{\alpha}}{\bm{\phi}(\vect{y}_j^\text{nav})}$
for some $\bm{\alpha} \in \Complex^{N_\ell}$.
The previous arguments are put together in
$y \approx \innerp{\bm{\Phi}(\vect{\Lambda})\bm{\alpha}}{\bm{\Phi}(\vect{\Lambda})\bm{\beta}_j} 
   = \bm{\alpha}^\hermconj\bm{\Phi}(\vect{\Lambda})^\hermconj\bm{\Phi}(\vect{\Lambda})\bm{\beta}_j
   = \bm{\alpha}^\hermconj\vect{K}\bm{\beta}_j 
~\text{s.t.}~\vect{1}_{N_\ell}^\top \bm{\beta}_j = 1$. The previous model
is summarized into the following hypothesis.

\begin{assumption}\label{as:data.approximation}
  There exist an $N_{\text{k}} \times N_\ell$ matrix $\vect{A}_1$, $N_\ell \times 
  N_\text{fr}$ sparse matrix $B$, and an
  $N_{\text{k}}\times N_{\text{fr}}$ matrix $\vect{E}_1$, gathering all approximation
  errors, s.t.\ $\vect{Y} = \vect{A}_1 {\vect{KB}} + \vect{E}_1$.
\end{assumption}

The dimensions of the kernel matrix $\vect{K}$ is contingent on the number of landmark 
points chosen and therefore, $\vect{K}$ can still be high dimensional depending on the 
physical characteristics of the acquired data. To impose a low-rank structure in As.~\ref{as:data.approximation}, and to meet the restrictions 
imposed by limited computational resources, dimensionality reduction on 
$\vect{K}$ is desirable. To this end, the methodology of~\cite{RSE.13} is 
applied:
\begin{subequations}\label{eq:LLE}
  \begin{enumerate}
    \item 
    Since $\{\bm{\phi}
    (\ell)_j\}_{j=1}^{N_\ell}$ lie on smooth manifold $\mathpzc{M}$, and similar to the earlier discussion, there exists a matrix 
    $\vect{W} \in \Complex^{N_{\ell}\times N_{\ell}}$ 
    s.t. $\bm{\Phi}(\vect{\Lambda}) \approx \bm{\Phi}
    (\vect{\Lambda})\vect{W}$, with $\vect{1}_{N_{\ell}}^{\intercal} 
    \vect{W} = \vect{1}_{N_{\ell}}^{\intercal}$ and $\diag(\vect{W})
              = \vect{0}$, where $\vect{1}_{N_{\ell}}^{\intercal} \vect{W} = \vect{1}_{N_{\ell}}^{\intercal}$
    manifests the desire for affine combinations and $\text{diag}(\vect{W}) = \vect{0}$
    is used to exclude the trivial solution of $\vect{W} = \vect{I}_{N_\ell}$. To use 
    the kernel trick, $\bm{\Phi}(\vect{\Lambda})^\hermconj
    \bm{\Phi}(\vect{\Lambda}) \approx \bm{\Phi}(\vect{\Lambda})^\hermconj\bm{\Phi}(\vect{\Lambda})\vect{W} \Rightarrow
    \vect{K} \approx \vect{KW}$. To summarize this step, given $\vect{K}$ and $\lambda_W > 0$, solve
        $
            \min_{\vect{W}}
            \norm*{\vect{K} - \vect{K}
            \vect{W}}_{\text{F}}^2 + \lambda_W \norm{\vect{W}}_1
            ~\text{s.to}\
            \vect{1}_{N_{\ell}}^{\intercal} \vect{W} =
              \vect{1}_{N_{\ell}}^{\intercal}\ \text{and}\ \diag(\vect{W})
              = \vect{0}
        $.
    This task is an affinely constrained convex optimization task and
    hence can be solved by~\cite{slavakis.FMHSDM}. 
    
    \item After $\vect{W}$ is obtained, and for a user-defined 
    integer $d \ll N_\ell$, solve
    $
        \min_{\check{\vect{K}}\in \Complex^{d\times
        N_{\ell}}}
        \norm{\check{\vect{K}} -
          \check{\vect{K}} \vect{W}}_{\text{F}}^2\ 
          ~\text{s.to}\
          \check{\vect{K}} \check{\vect{K}}^{\hermconj} = \vect{I}_d
    $ to reduce the dimension of $\vect{K}$ to $d$ while preserving the 
    underlying manifold geometry in the original high-dimensional space. 
    The constraint $\check{\vect{K}} \check{\vect{K}}^{\hermconj} = \vect{I}_d$
    is used to exclude the trivial solution of $\check{\vect{K}} = \vect{0}$. The solution of the previous task is the Hermitian 
    transpose of the matrix comprising the $d$ minimal
    eigen-vectors of $(\vect{I}_{N_{\ell}} - \vect{W}) (\vect{I}_{N_{\ell}} - \vect{W})^{\hermconj}$.
\end{enumerate}
\end{subequations}
Other methods for reducing the dimension of $\vect{K}$, such as extracting a random subset
of its rows, are reserved for a future journal publication due to 
lack of space. The previous discussion raises the need to unfold the kernel 
matrix $\check{\vect{K}}$ to its original dimensions. 
The following ``decompression'' hypothesis establishes a linear relation 
between $\vect{K}$ and $\check{\vect{K}}$.
\begin{assumption}\label{as:ker.dimension.reduction}
  There exist an $N_{\ell} \times d$ matrix $\vect{A}_2$ and an
  $N_{\ell}\times N_{\ell}$ matrix $\vect{E}_2$, gathering all approximation
  errors, s.t.\ $\vect{K} = \vect{A}_2 \check{\vect{K}} + \vect{E}_2$.
\end{assumption}

\section{The Non-Convex Inverse Problem and its Algorithmic Solution}\label{sec:reconstruction}

By As.~\ref{as:data.approximation} and \ref{as:ker.dimension.reduction}, the (k,t)-space 
measurements $\vect{Y}$ can be modeled as $\vect{Y} = \vect{A}\check{\vect{K}}\vect{B} + \vect{E}$.
Upon defining, $\vect{D} := \mathcal{F}^{-1}(\vect{A})$, $\vect{A}\check{\vect{K}}\vect{B} = 
\mathcal{F}(\vect{D})\check{\vect{K}}\vect{B}$, and by the virtue of the linearity of 
$\mathcal{F}$, $\vect{A\check{K}B} = \mathcal{F}(\vect{D}\check{\vect{K}}\vect{B})$. This
establishes the following bi-linear relation between $\vect{Y}$ and the unknowns 
$(\vect{D}, \vect{B})$: $\vect{Y} = \mathcal{F}(\vect{D}\check{\vect{K}}\vect{B}) + \vect{E}$. 
This bi-linear 
relation holds true also in the image domain: $\vect{X} =
\mathcal{F}^{-1}(\vect{Y}) = \vect{D}\check{\vect{K}}\vect{B} + \mathcal{F}^{-1}(\vect{E})$.

As discussed in Sec.~\ref{sec:data.rep}, only a few measurements are available in the 
(k,t)-space. To account for missing entries, a sampling operator $\mathcal{S}(\cdot)$
is introduced to emulate sampling trajectories. The operation $\mathcal{S}(\vect{Y})$ 
retrospectively under-samples $\vect{Y}$ retaining the sampled k-space locations 
and filling the other locations with zeros. Given the positive real-valued parameters
$(\lambda_1, \lambda_2, \lambda_3, C_D)$, the following inverse problem is 
formulated:
\begin{align}
  \min_{(\vect{D}, \vect{B}, \vect{Z})} 
  &\ \overbrace{\tfrac{1}{2} \norm*{\mathcal{S}(\vect{Y}) -
    \mathcal{S} \mathcal{F}(\vect{D} 
    \check{\vect{K}} \vect{B})}_{\text{F}}^2}^{\text{T1}} 
    + \overbrace{\tfrac{\lambda_1}{2} \norm*{\vect{Z} -
    \mathcal{F}_t(\vect{D} \check{\vect{K}}
    \vect{B})}_{\text{F}}^2}^{\text{T2}}
  \ + \overbrace{\lambda_2 \norm*{\vect{B}}_1}^{\text{T3}} + \overbrace{ \lambda_3
    \norm*{\vect{Z}}_1}^{\text{T4}} \notag \\
  &\ \text{s.to}\
  \underbrace{\norm*{\vect{De}_i} \leq C_D,\ \forall i\in\Set{1, \ldots,
    d}}_{\text{C1}};~\underbrace{\vect{1}_{N_{\ell}}^{\intercal} \vect{B} = 
    \vect{1}_{N_{\text{fr}}}^{\intercal}}_{\text{C2}};
  \vect{D}\in \Complex^{N_{\text{k}} \times d}; \vect{B}\in \Complex^{N_{\ell}
    \times N_{\text{fr}}}; \vect{Z}\in\Complex^{N_{\text{k}} \times
    N_{\text{fr}}} , 
    \label{eq:recovery.task} 
\end{align}
where $\vect{e}_i$ denotes the $i\text{th}$ column of the identity matrix $\vect{I}_d$.
In addition to the data fit term T1,~\eqref{eq:recovery.task} 
contains terms T2 and T4 where the auxiliary variable $\vect{Z}$ is introduced 
to capture the periodic process, (\eg beating heart motion) by 
imposing sparsity on $\mathcal{F}_t(\vect{D}\check{\vect{K}}\vect{B})$. 
$\mathcal{F}_t(\cdot)$ performs 1D Fourier Transform of the 
$1 \times \text{N}_\text{fr}$ time profile of every single pixel. The term T3 imposes 
the sparsity constraint and C2 accounts for the affine constraint 
on $\vect{B}$ as in As.~\ref{as:data.approximation}. Term C1 prevents the unbounded solution 
of $\vect{D}$ due to the scaling ambiguity in $\vect{D}\check{\vect{K}}\vect{B}$.

To solve~\eqref{eq:recovery.task}, the successive-convex-approximation framework of 
~\cite{facchinei2015parallel} is employed and is presented in 
Alg.~\ref{alg:recovery}. Convergence to a stationary solution of~\eqref{eq:recovery.task}
is guaranteed~\cite{facchinei2015parallel}. Step~\ref{alg.step:convex.tasks} of 
Alg.~\ref{alg:recovery} calls for solving convex minimization sub-tasks.  
Given $\tau_U, \tau_B > 0$, for $(\vect{D}_n, \vect{B}_n, \vect{Z}_n)$ at every
iteration of the algorithm, the following estimates are required:
\begin{subequations}\label{eq:solveDBZ}
  \begin{align}
    &\ \hat{\vect{D}}_n\in
    \Argmin_{\vect{D}}\,
    \tfrac{1}{2} \norm*{\mathcal{S}(\vect{Y}) -
      \mathcal{S} 
      \mathcal{F}(\vect{D} \check{\vect{K}}
      \vect{B}_n)}_{\text{F}}^2 + \tfrac{\tau_D}{2}
      \norm*{\vect{D} - \vect{D}_n}_{\text{F}}^2
    + \tfrac{\lambda_1}{2}
      \norm*{\vect{Z}_n - \mathcal{F}_t(\vect{D} \check{\vect{K}} 
      \vect{B}_n)}_{\text{F}}^2
    \ \text{s.to}\ \norm*{\vect{De}_i} \leq C_D,\
                   \forall i\in\Set{1, \ldots, d}\,. \label{eq:min.for.D} \\
    &\ \hat{\vect{B}}_n\in \Argmin_{\vect{B}} 
    \tfrac{1}{2} \norm*{\mathcal{S}(\vect{Y}) - \mathcal{S} \mathcal{F}(\vect{D}_n
      \check{\vect{K}} \vect{B})}_{\text{F}}^2 + \tfrac{\tau_B}{2} \norm*{\vect{B}-
      \vect{B}_n}_{\text{F}}^2
    + \tfrac{\lambda_1}{2} \norm*{\vect{Z}_n - \mathcal{F}_t(\vect{D}_n \check{\vect{K}}
      \vect{B})}_{\text{F}}^2 + \lambda_3\norm*{\vect{B}}_1
    \ \text{s.to}\
    \vect{1}_{N_{\ell}}^{\intercal} \vect{B} =
      \vect{1}_{N_{\text{fr}}}^{\intercal} \,. \label{eq:min.for.B}
  \end{align}
\end{subequations}

Tasks in~\eqref{eq:solveDBZ} can be viewed as affinely constrained composite convex
minimization tasks, hence allowing the use of~\cite{slavakis.FMHSDM}. The algorithmic 
and mathematical details of~\eqref{eq:solveDBZ} are reserved for 
an upcoming journal publication.

\begin{algorithm}[!t]
  \begin{algorithmic}[1]
    \renewcommand{\algorithmicindent}{1em}
    \addtolength\abovedisplayskip{-0.5\baselineskip}%
    \addtolength\belowdisplayskip{-0.5\baselineskip}%

    \Require{Available are $\mathcal{S}(\vect{Y})$ and navigator data
      $\vect{Y}_{\text{nav}}$. Choose parameters
      $\lambda_1, \lambda_2, \lambda_3, C_D, \tau_D, \tau_B >0$, as well as
      $\zeta\in (0,1)$ and $\gamma_0\in (0,1]$.}

    \Ensure{Extract the limit points $\vect{D}_*$ and $\vect{B}_*$ of sequences
      $(\vect{D}_n)_n$ and $(\vect{B}_n)_n$, respectively, and recover the dMRI data
      as $\hat{\vect{X}} := \vect{D}_* \check{\vect{K}}\vect{B}_*$.}

    \State\parbox[t]{\dimexpr\linewidth-\algorithmicindent}%
    {Identify landmark points $\bm{\Lambda}$ from the columns of
      $\vect{Y}_{\text{nav}}$ according to~\cite{Landmark.MinMax}.}
    
    \State\parbox[t]{\dimexpr\linewidth-\algorithmicindent}%
    {Compute the kernel matrix $\vect{K}$ from landmark points
    $\bm{\Lambda}$.}
    
    \State\parbox[t]{\dimexpr\linewidth-\algorithmicindent}%
    {Compute the ``compressed'' $\check{\vect{K}}$ (\cf~Sec.~\ref{sec:background}).}

    \State\parbox[t]{\dimexpr\linewidth-\algorithmicindent}%
    {Arbitrarily fix $(\vect{D}_0, \vect{B}_0, \vect{Z}_0)$ and set $n=0$.}

    \While{$n\geq 0$}\label{alg.step:resume}

    \State\parbox[t]{\dimexpr\linewidth-\algorithmicindent}%
    {Available are $(\vect{D}_n, \vect{B}_n, \vect{Z}_n)$ and
      $\gamma_n$.}

    \State\parbox[t]{\dimexpr\linewidth-\algorithmicindent}%
    {Let $\gamma_{n+1} := \gamma_n (1-\zeta\gamma_n)$.}
    
    \State\parbox[t]{\dimexpr\linewidth-\algorithmicindent}%
    {Obtain $\hat{\vect{D}}_n$ of \eqref{eq:min.for.D} and $\hat{\vect{B}}_n$ of
      \eqref{eq:min.for.B}, and the $(i,j)$th entry of
      $\hat{\vect{Z}}_n$, $\forall (i,j)$, via 
      $\vect{A}_{\text{aux}} \coloneqq \mathcal{F}_t(\vect{D}_n \check{\vect{K}} \vect{B}_n)$ and the following soft-thresholding rule:
      \begin{align*}
        [\hat{\vect{Z}}_n]_{ij}\ \mathbin{:=}\
        [\vect{A}_{\text{aux}}]_{ij}
        \Biggl(1 -
        \frac{\lambda_3/\lambda_1}{\max \left\{\lambda_3/ \lambda_1,
        \left\lvert [\vect{A}_{\text{aux}}]_{ij} \right\rvert
        \right\}} \Biggr) \,.
      \end{align*}}\label{alg.step:convex.tasks}

    \State\parbox[t]{\dimexpr\linewidth-\algorithmicindent}%
    {Update
      \begin{align*}
        (\vect{D}_{n+1}, \vect{B}_{n+1}, \vect{Z}_{n+1})\
        \mathbin{:=}\
        (1-\gamma_{n+1}) (\vect{D}_n, \vect{B}_n,
          \vect{Z}_n)
        + \gamma_{n+1} (\hat{\vect{D}}_{n},
        \hat{\vect{B}}_{n}, \hat{\vect{Z}}_{n}) \,.
      \end{align*}
    }

    \State\parbox[t]{\dimexpr\linewidth-\algorithmicindent}%
    {Set $n$ equal to $n+1$ and go to step~\ref{alg.step:resume}.}

    \EndWhile\label{alg.step:endwhile}
    
  \end{algorithmic}
  \caption{Recovering dMRI data}\label{alg:recovery}
\end{algorithm}

\section{Numerical Results}\label{sec:numerical}

The proposed framework is validated on the magnetic-resonance
extended cardiac-torso (MRXCAT) cine phantom~\cite{wissmann2014mrxcat}. Further 
investigation via testing on other datasets is beyond the scope 
of this paper due to lack of space.  The extended cardiac torso (XCAT) framework 
was used to generate a breath-hold 
cardiac cine data of spatial size $(N_\text{p}, N_\text{f}) = (408, 408)$ with 
$N_\text{fr} = 360$ number of frames. The generated data has a spatial resolution of 
$1.56 \times 1.56~\text{mm}^2$ for a field-of-view (FOV) of $400 \times 400~\text{mm}^2$. 
The phantom data consists of $15$ cardiac cycles and $24$ cardiac phases. The data is
retrospectively undersampled under 1D Cartesian sampling strategy, as in Fig.~\ref{fig:sampling.strategy.cartesian}, to emulate an undersampling rate
[defined by $N_\text{k}N_\text{fr}/ (\text{\# of acquired voxels})$] of $8$. 
Normalised root mean square error (NRMSE), defined as $\|\vect{X} - \vect{\hat{X}}\|_\text{F}/\|\vect{X}\|_\text{F}$, is used to measure the quality of 
reconstruction, with $\vect{X}$ being the 
fully sampled, high-fidelity, original MR image and $\vect{\hat{X}}$ the 
estimate of $\vect{X}$ obtained from various reconstruction schemes.

\begin{figure*}[t]
    \centering
    \subfloat{\begin{annotate}
        {\includegraphics[width=0.15\textwidth, trim={110 60 110 40}, clip]
        {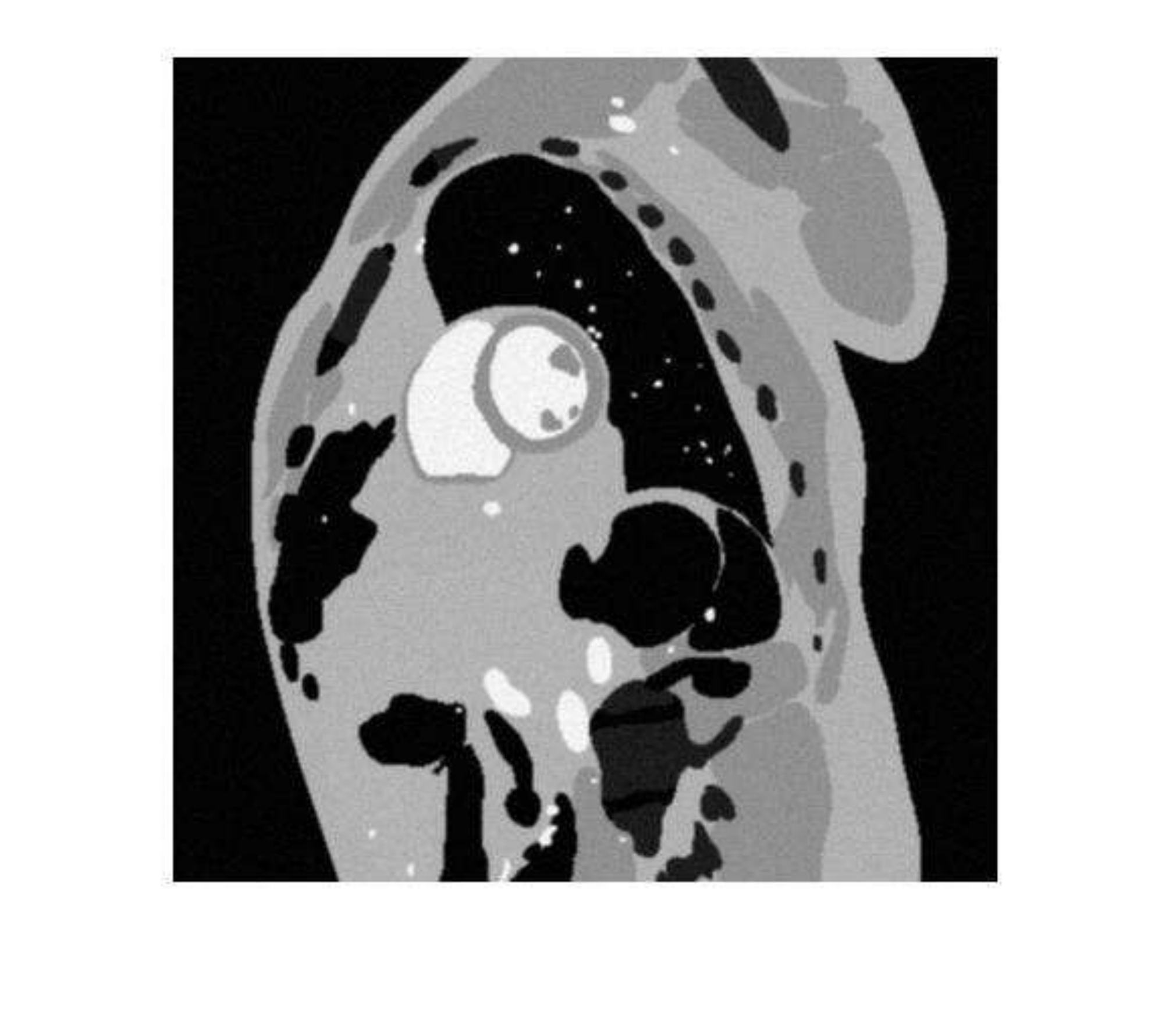}}{0.5}
        \end{annotate}}
    \subfloat{\begin{annotate}
        {\hspace{-0.5cm}
        \includegraphics[width=0.15\textwidth, trim={110 60 110 40}, clip]
        {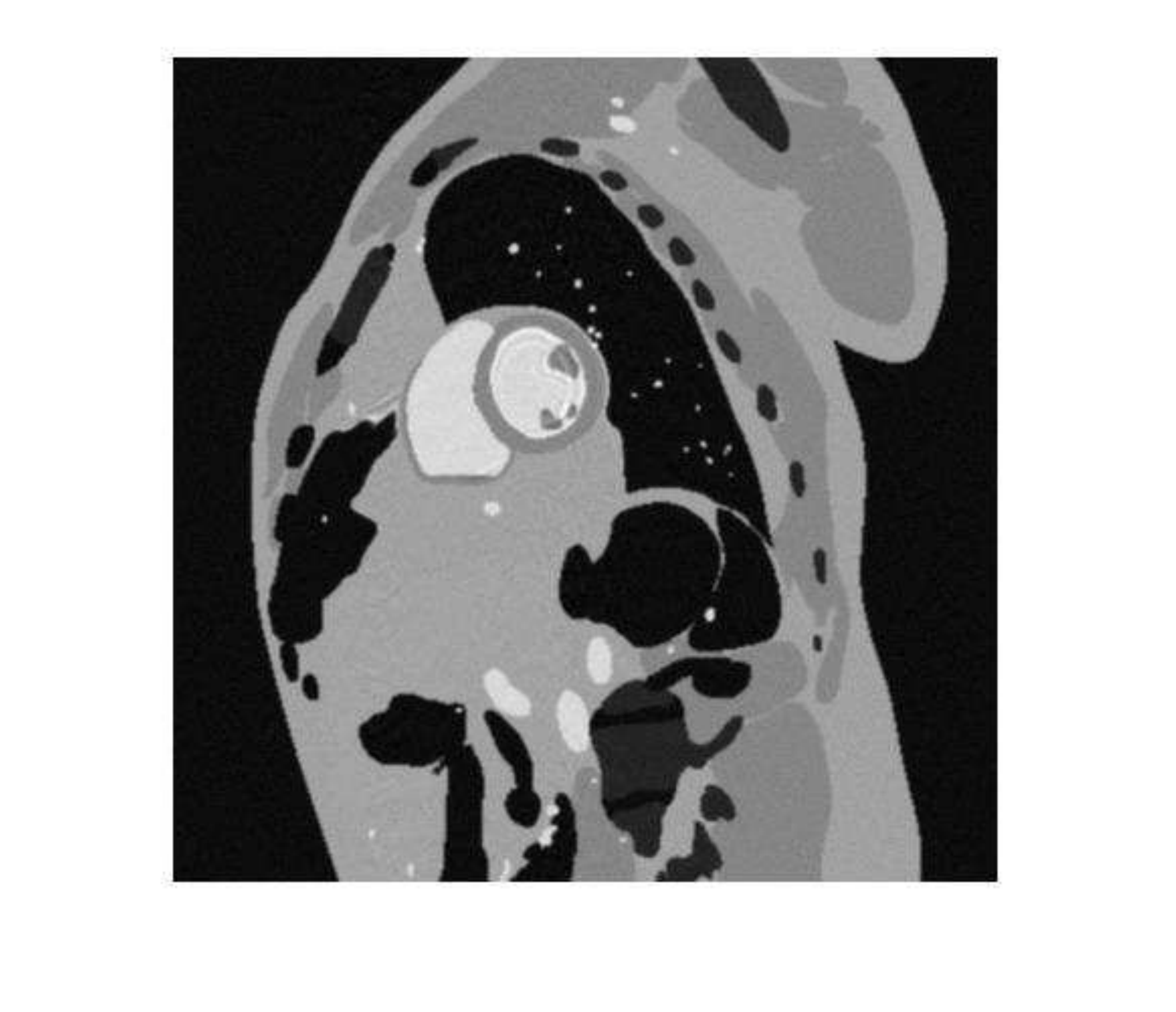}}{0.5}
        \draw[red, ->, thick] (-1.5,1.8) -- (-1, 1);
        \end{annotate}}
    \subfloat{\begin{annotate}
        {\hspace{-0.5cm}
        \includegraphics[width=0.15\textwidth, trim={110 60 110 40}, clip]
        {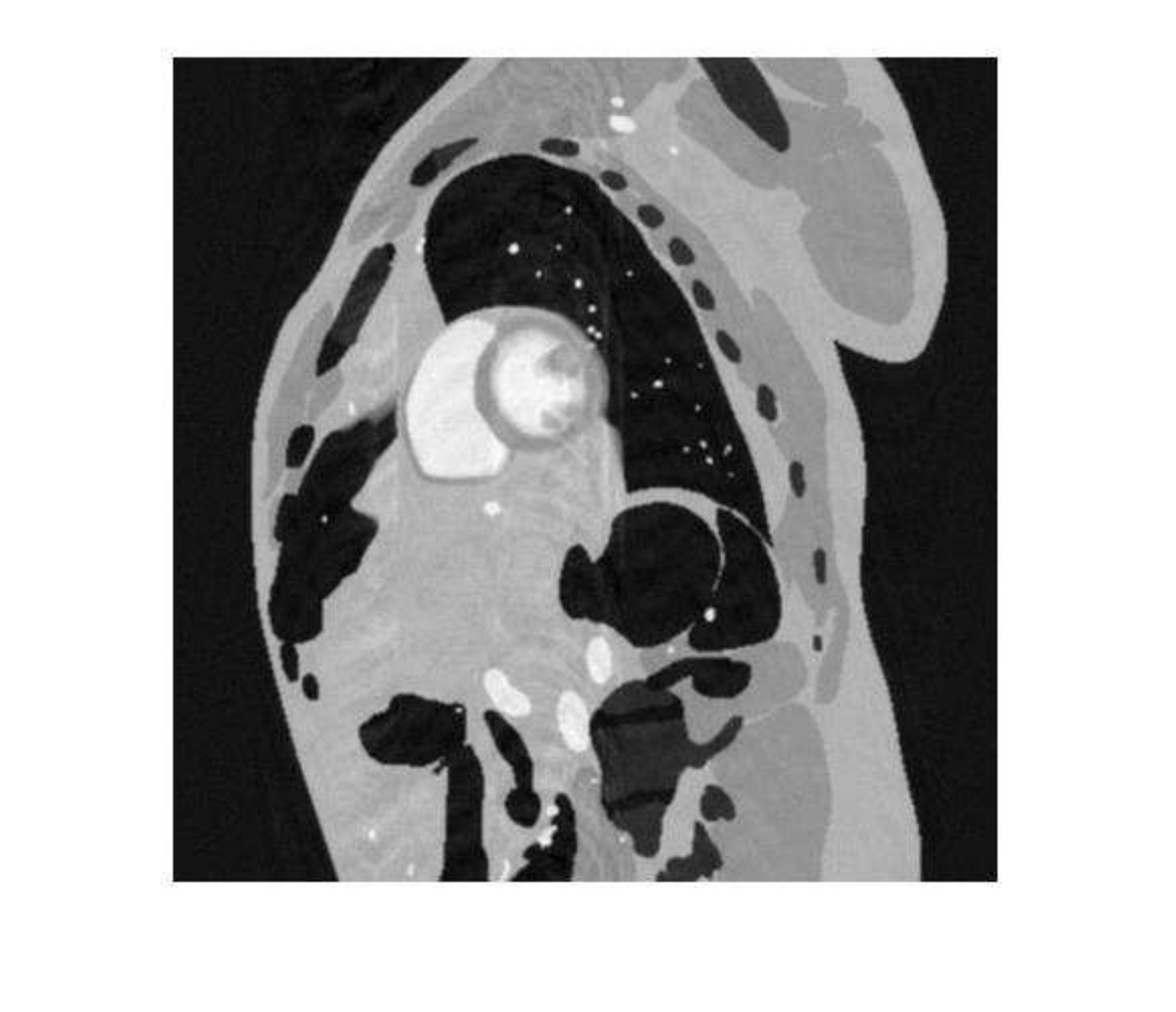}}{0.5}
        \draw[red, ->, thick] (-1.5,1.8) -- (-1, 1);
        \end{annotate}}
    \subfloat{\begin{annotate}
        {\hspace{-0.5cm}
        \includegraphics[width=0.15\textwidth, trim={110 60 110 40}, clip]
        {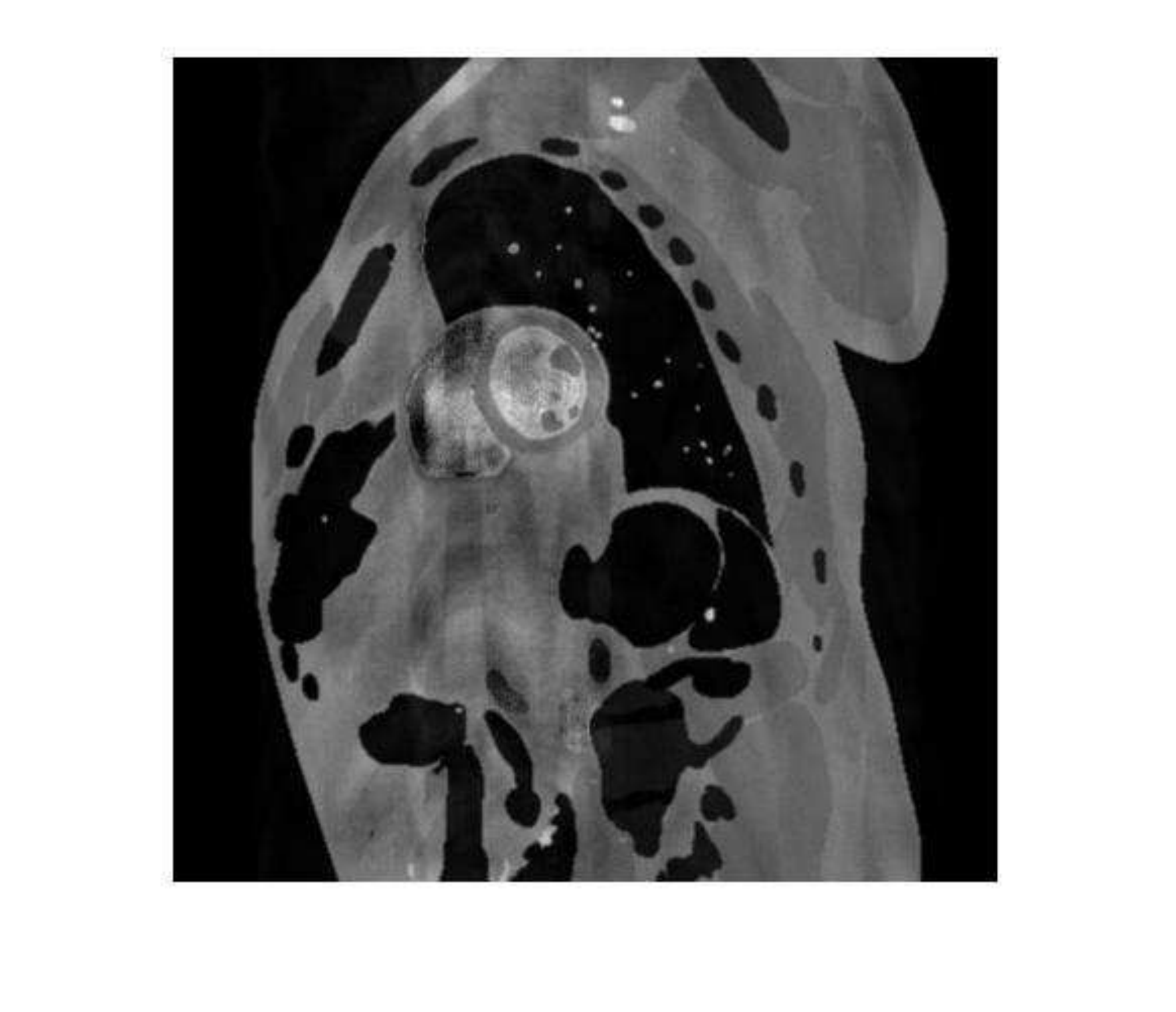}}{0.5}
        \draw[yellow, thick] (-1.8,1.4) rectangle (0,0);
        \end{annotate}}
    \subfloat{\begin{annotate}
        {\hspace{-0.5cm}
        \includegraphics[width=0.15\textwidth, trim={110 60 110 40}, clip]
        {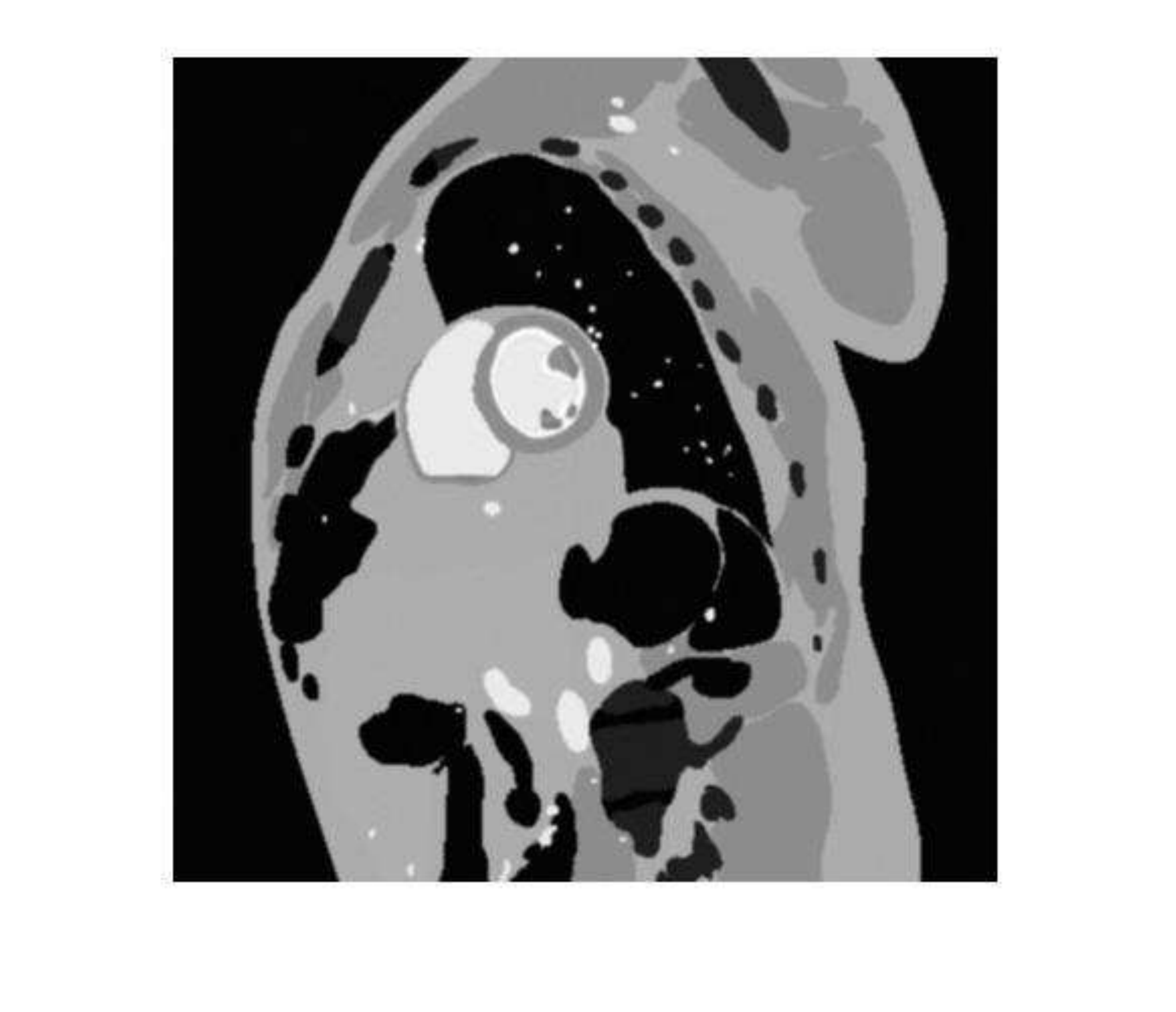}}{0.5}
        \end{annotate}}
    \subfloat{\begin{annotate}
        {\hspace{-0.5cm}
        \includegraphics[width=0.15\textwidth, trim={110 60 110 40}, clip]
        {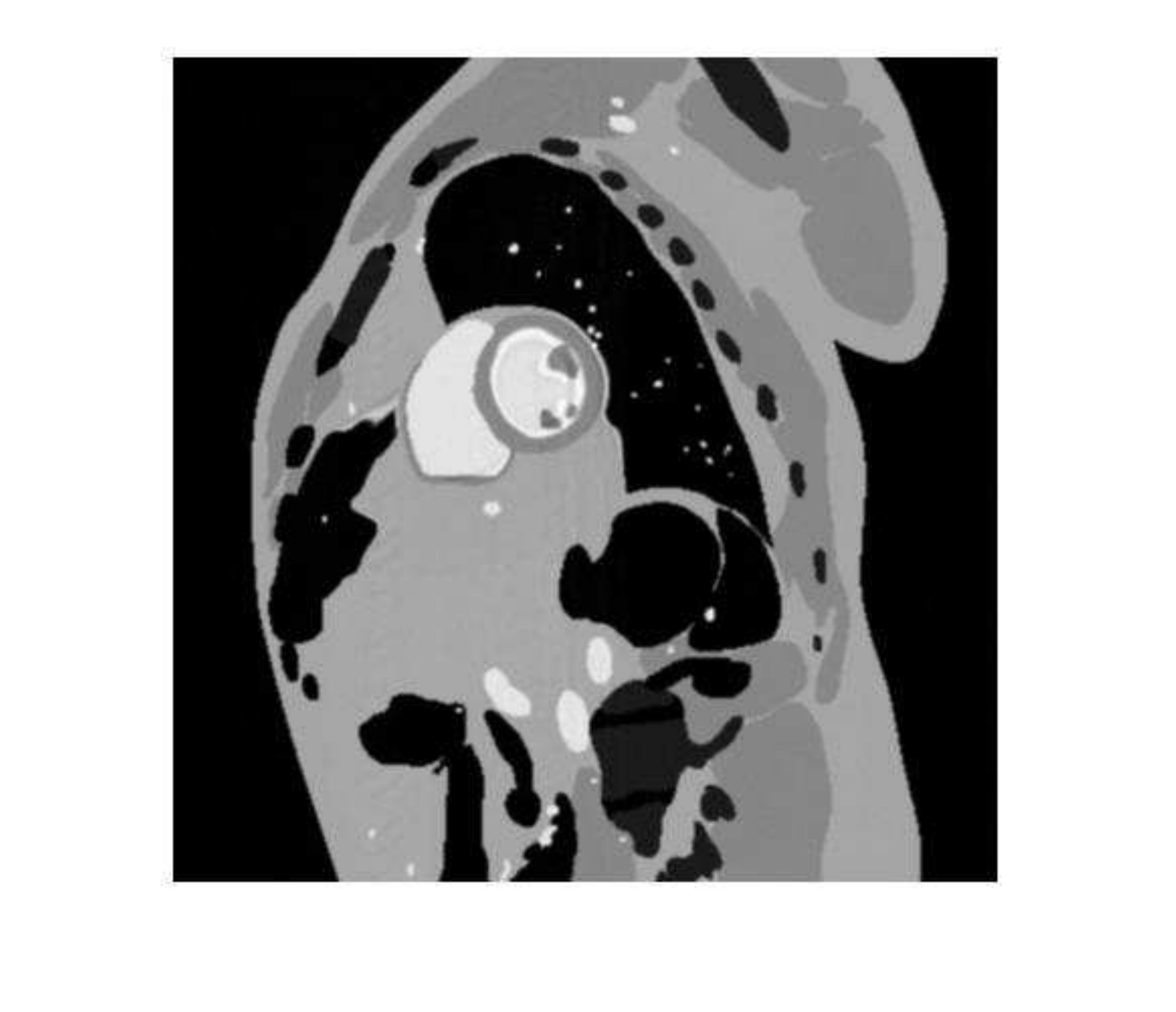}}{0.5}
        \draw[yellow, thick] (-1.8,1.4) rectangle (0,0);
        \end{annotate}} \\
        \vspace{-0.5cm}
    \subfloat{\begin{annotate}
        {\includegraphics[width=0.15\textwidth, trim={110 60 110 40}, clip]
        {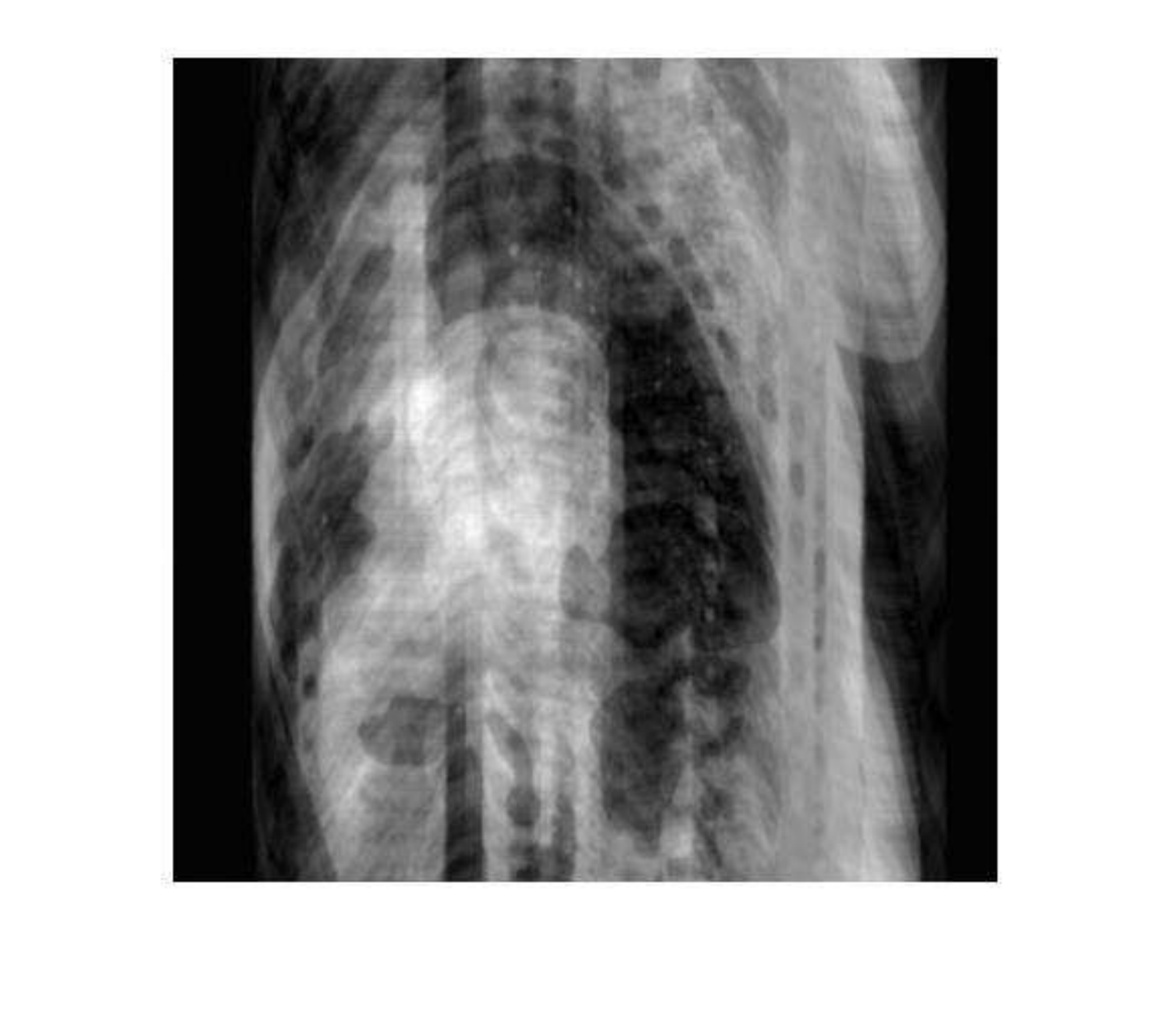}}{0.5}
        \end{annotate}}
    \subfloat{\begin{annotate}
        {\hspace{-0.5cm}
        \includegraphics[width=0.15\textwidth, height=3.03cm]
        {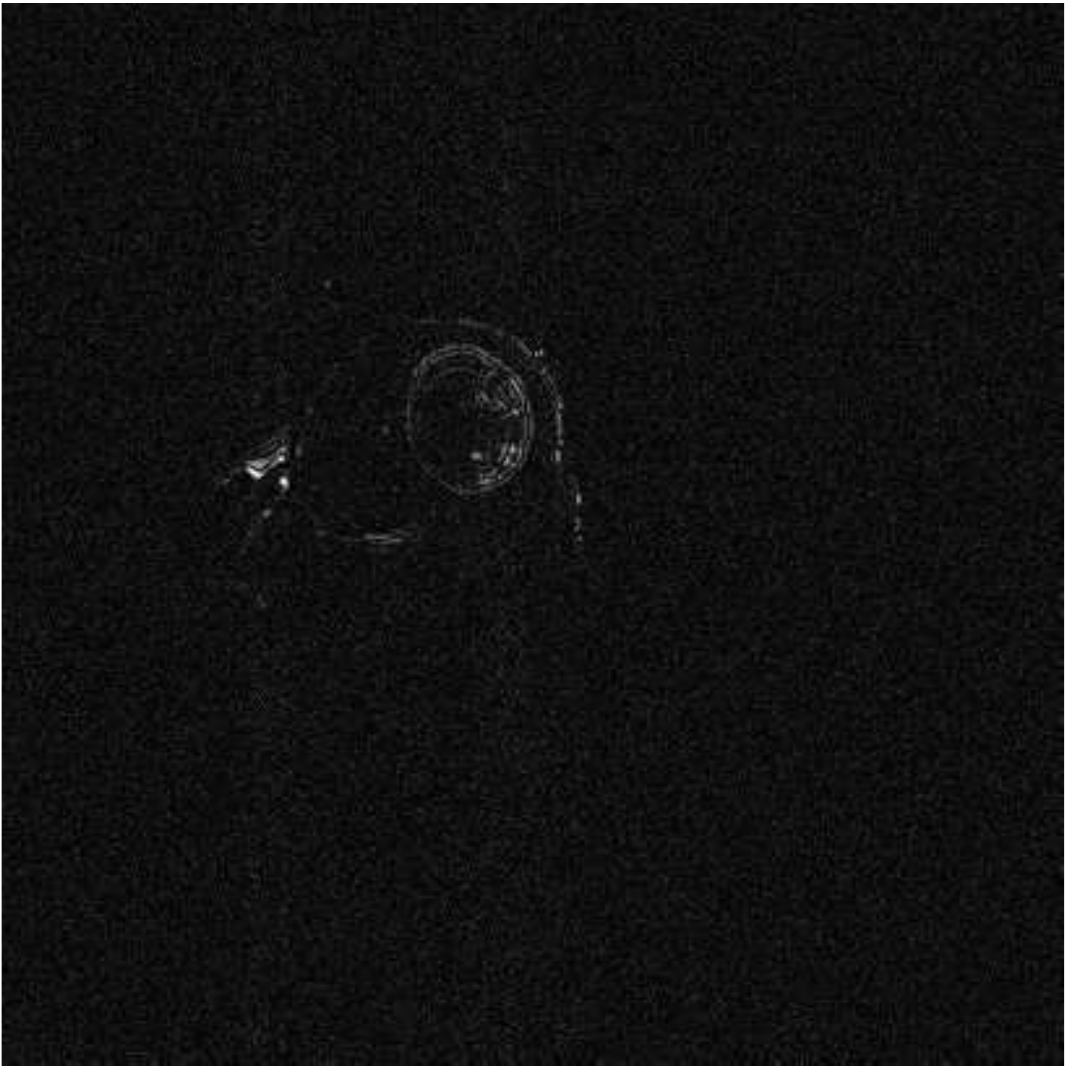}}{0.5}
        \end{annotate}}
    \subfloat{\begin{annotate}
        {\hspace{-0.5cm}
        \includegraphics[width=0.15\textwidth, height=3.03cm]
        {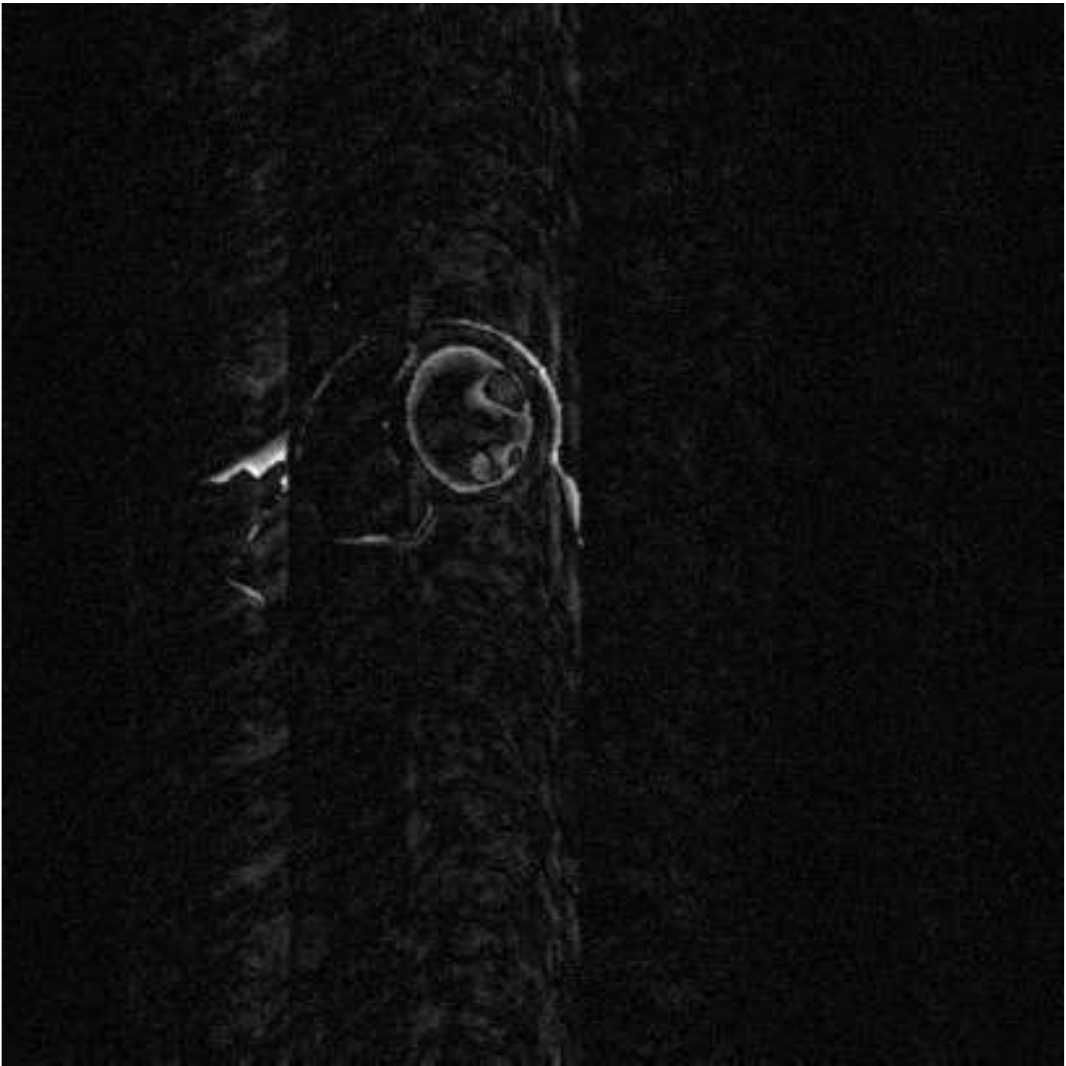}}{0.5}
        \end{annotate}}
    \subfloat{\begin{annotate}
        {\hspace{-0.5cm}
        \includegraphics[width=0.15\textwidth, height=3.03cm]
        {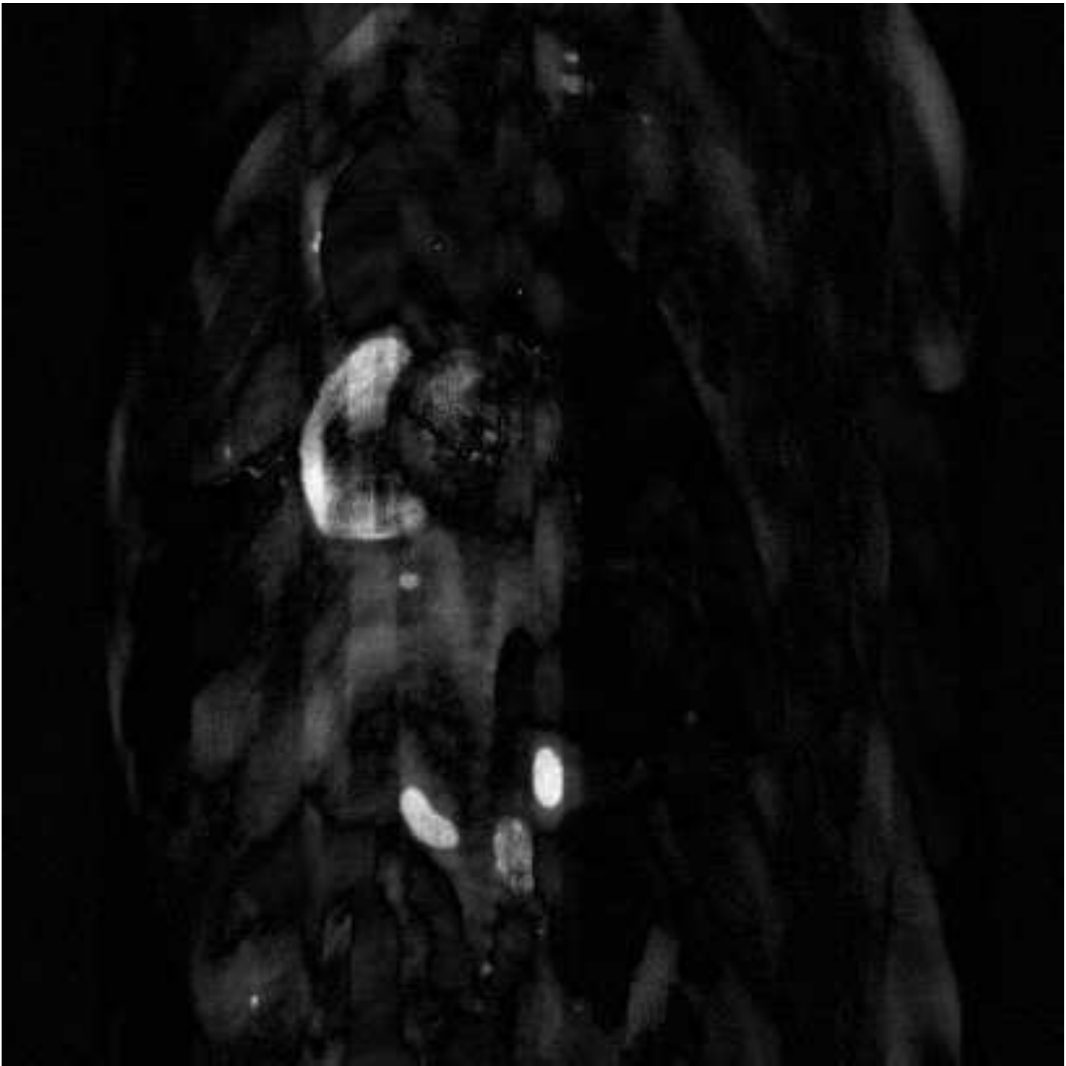}}{0.5}
        \end{annotate}}
    \subfloat{\begin{annotate}
        {\hspace{-0.5cm}
        \includegraphics[width=0.15\textwidth, height=3.03cm]
        {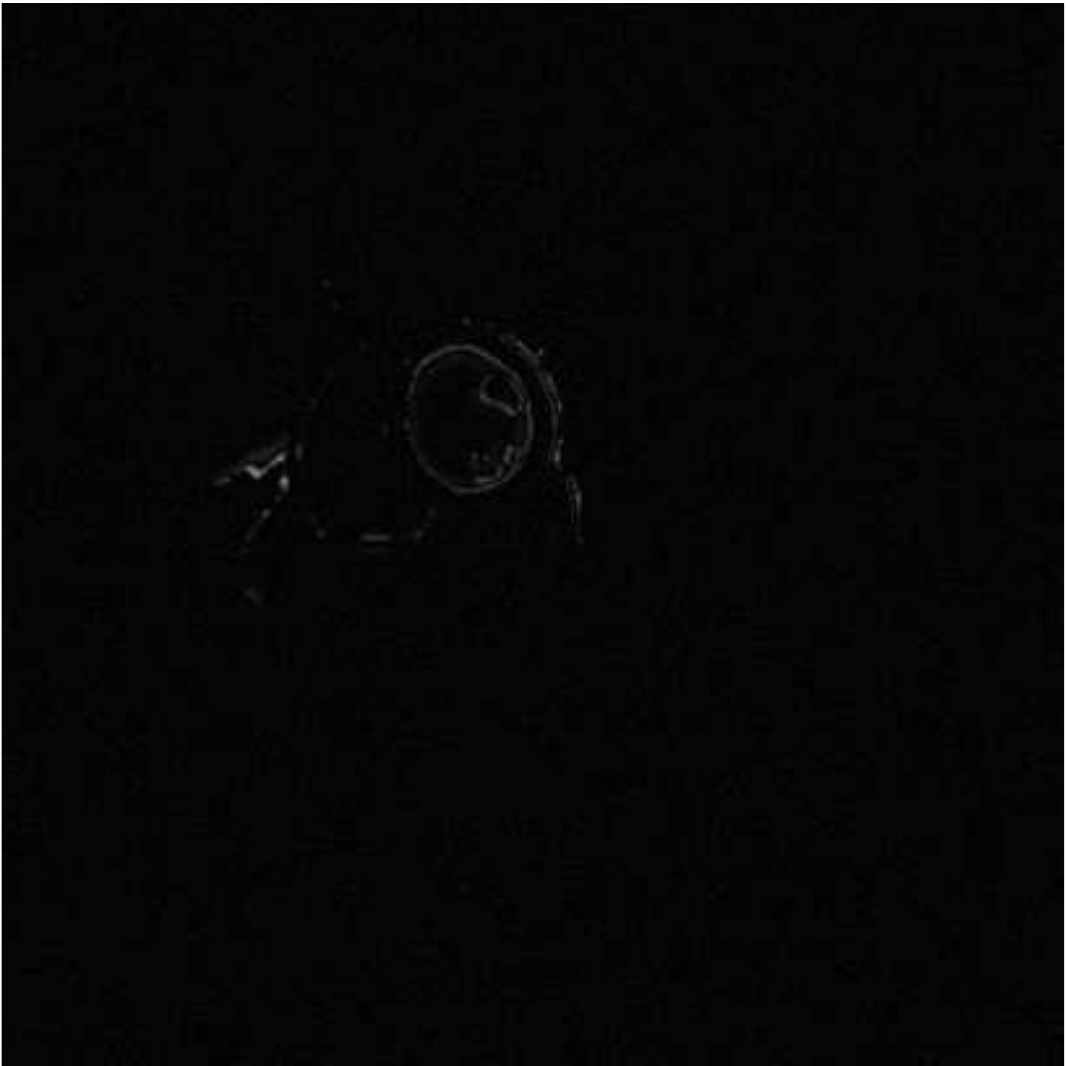}}{0.5}
        \draw[yellow, ->, thick] (-1.5,1.8) -- (-1, 1);
        \end{annotate}}
    \subfloat{\begin{annotate}
        {\hspace{-0.5cm}
        \includegraphics[width=0.15\textwidth, height=3.03cm]
        {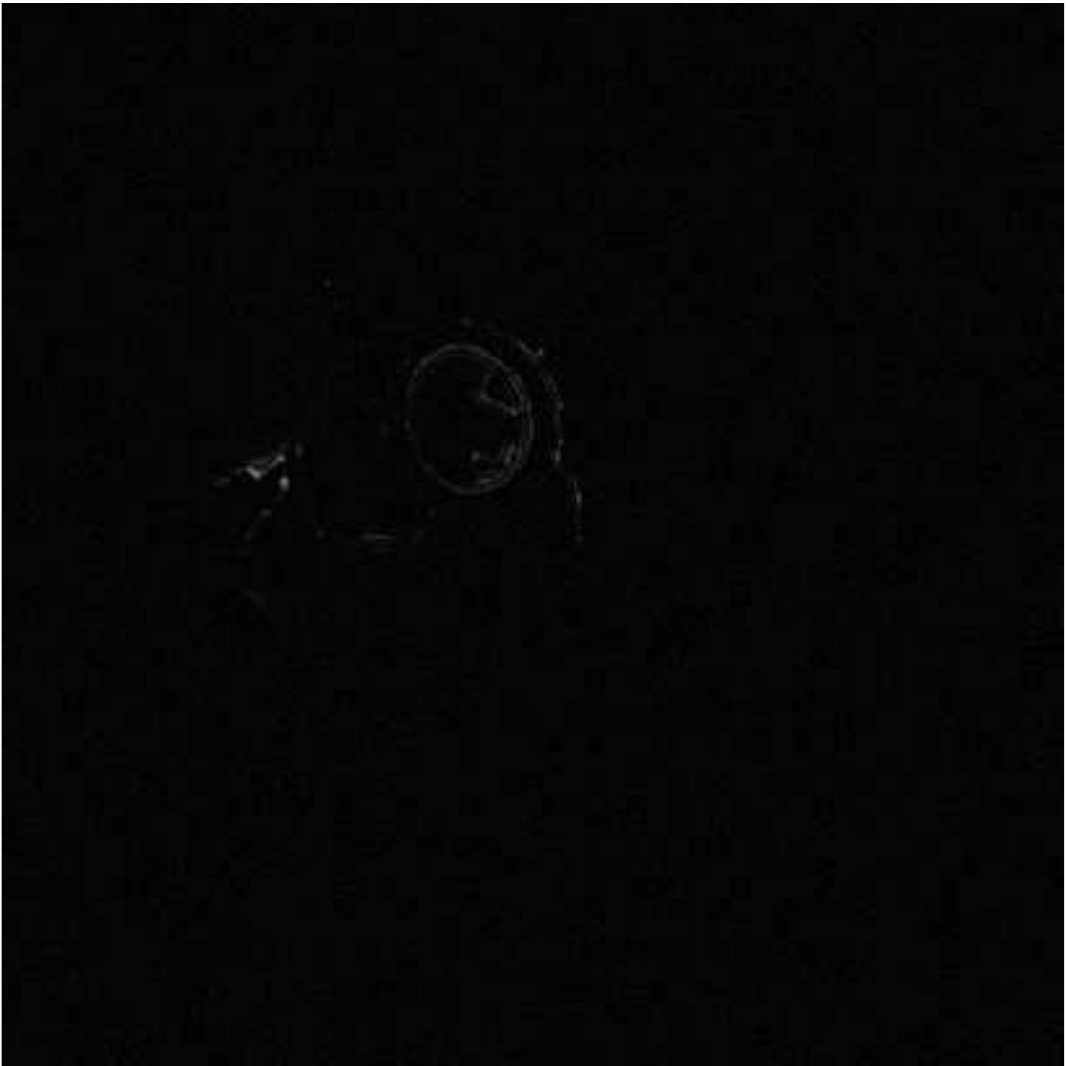}}{0.5}
        \draw[yellow, ->, thick] (-1.5,1.8) -- (-1, 1);
        \end{annotate}}
    \caption{Spatial Results for frame 49 of MRXCAT cardiac cine phantom
    retrospectively undersampled at acceleration rate of 8. 
    Left to Right: Gold standard (top row); Undersampled Image (bottom row), 
    PS-Sparse ($0.0556$), SToRM ($0.0503$), KLR ($0.0744$), BiLMDM ($0.0456$), 
    KBiLMDM ($0.0389$). The numerical values indicate the NRMSE. 
    Top row: Spatial frames generated from aforementioned reconstructed 
    schemes; bottom row: Corressponding error maps.}
    \label{fig:mrxcat.f49}
\end{figure*}

\begin{figure}[!t]
    \centering
    \includegraphics[width=0.5\linewidth, trim={32 10 55 25}, clip]
    {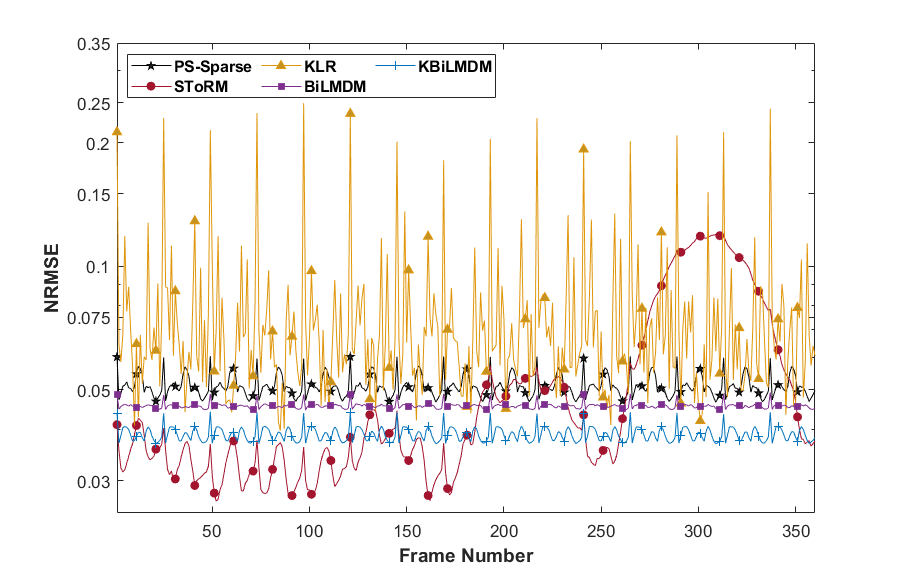}
    \caption{Frame-wise NRMSE of PS-Sparse $(0.056 \pm 3.2\times10^{-3})$, 
    SToRM $(0.0521 \pm 2.6\times10^{-2})$, 
    KLR $(0.0753 \pm 3.6\times10^{-2})$, BiLMDM $(0.0461 \pm 7.6\times10^{-4})$,
    KiBLMDM $(0.0395 \pm 1.6\times10^{-3})$. The
    previous numerical values indicate mean and standard deviation 
    of the NRMSE across 360 frames.}
    \label{fig:framewise.error}
\end{figure}

The performance of KBiLMDM is tested against various state-of-the-art techniques,
namely, the partially separable sparsity aware model (PS-Sparse)~\cite{zhao2012pssparse}, 
smoothness regularization on manifolds (SToRM)~\cite{poddar2016dynamic}, 
kernel low rank (KLR)~\cite{nakarmi2017kernel}, and the bi-linear modeling of data manifolds (BiLMDM)~\cite{shetty2019bi}. KBiLMDM achieves the least NRMSE value of $0.03891$ as compared 
to PS-Sparse's $0.0556$, SToRM's $0.0503$,
KLR's $0.0744$, BiLMDM's $0.0456$. KBiLMDM shows less deviation in NRMSE than KLR, SToRM and
PS-Sparse while consistently scoring the least NRMSE as 
compared to BiLMDM, KLR and PS-Sparse across all frames. A further qualitative
analysis of the reconstructions, as in Fig.~\ref{fig:mrxcat.f49} provides 
evidence that KBiLMDM outperforms the other aforementioned state-of-the-art techniques.
KLR and SToRM exhibit severe aliasing effects as marked in 
Fig.~\ref{fig:mrxcat.f49}, and the error maps of PS-Sparse provides evidence that
reconstructions along the edges are not upto the mark, while there are discrepancies in 
the contrast as compared to the ground truth. On the other hand, KBiLMDM
produces sharper, aliasing-free, distortion-free dMRI time series. A comparison
of the error maps between BiLMDM and KBiLMDM shows improvements in the reconstructions
due to the introduction of kernels. Details on choice of kernels and 
{comparison with schemes like FRIST-MRI}~\cite{wen2017frist} are 
left for presentation and upcoming journal publications.

\section{Conclusion \& the Road Ahead}\label{sec:conclusion}


A kernel-based framework for reconstructing data on manifolds was
proposed and tailored to fit the dMRI-data recovery
problem. The proposed methodology exploited simple tangent-space
geometries of manifolds and followed classical kernel-approximation arguments to form a
bi-linear inverse estimation problem. Departing from
state-of-the-art approaches,
\begin{enumerate*}[label=\textbf{\roman*})]
\item no training data were used;
\item no graph Laplacian matrix was employed to penalize the inverse problem;
\item no costly (kernel) preimaging step was necessary to map feature points back to the input space; and
\item complex-valued kernel functions were defined to account for k-space data.
\end{enumerate*}
Preliminary numerical tests and qualitative investigation of
reconstructed images on retrospectively undersampled synthetic MR
images showed the rich potential of the proposed framework against
several state-of-the-art techniques. With regards to the road
ahead, the incorporation of training data into the framework is
currently under investigation. On-going research includes also
modeling extensions, further numerical tests on
a wide range of synthetic and real dMRI data,
which will be reported during the conference.

 


\end{document}